# Deep Control - a simple automatic gain control for memory efficient and high performance training of deep convolutional neural networks

Brendan Ruff [brendan@deepsee.ai]

**Abstract**

Training a deep convolutional neural net typically starts with a random initialisation of all filters in all layers which severely reduces the forward signal and back-propagated error and leads to slow and sub-optimal training. Techniques that counter that focus on either increasing the signal or increasing the gradients adaptively but the model behaves very differently at the beginning of training compared to later when stable pathways through the net have been established. To compound this problem the effective minibatch size varies greatly between layers at different depths and between individual filters as activation sparsity typically increases with depth leading to a reduction in effective learning rate since gradients may superpose rather than add and this further compounds the covariate shift problem as deeper neurons are less able to adapt to upstream shift.

Proposed here is a method of automatic gain control of the signal built into each convolutional neuron that achieves equivalent or superior performance than batch normalisation and is compatible with single sample or minibatch gradient descent. The same model is used both for training and inference.

The technique comprises a scaled per sample map mean subtraction from the raw convolutional filter output followed by scaling of the difference.

## 1 Introduction

Convolutional neural networks have become the workhorse of semantic interpretation and labelling of images, video, and other scene related signals such as depth and motion as they learn complex nonlinear rules without human intervention whatever the level of supervision. However they remain a challenge to train quickly even with high power parallel computation technology such as GPU. Not only does training incur a significant time delay before results are available but also consumes electricity and expensive computational resources which becomes ever more significant with the move to AI in cloud based services that scales costs hugely.

Consequently techniques have emerged to speed up the training process such as *batch normalisation* [2] that seeks to balance the signal distribution throughout the network, adaptive gradient balancing such as ADAM [3] that seeks to balance the gradient signal directly, and weight normalisation [4] that seeks to balance the weights in the convolutional filters to balance the signal strength.

The two main problems in training a deep convolutional neural network are *vanishing gradients* [5] and *covariate shift* [6], and it is worth to reconsider these now in light of improvements in deep network design. *Exploding gradients* is less of a problem and will not be discussed further here. The mathematical arguments to explain these problems will not be presented here, and in short vanishing gradients relates to severe attenuation of the error signal during backpropagation [7] as this error signal is modulated both by the forward signal to determine gradients for weight update and the by the gradient through the nonlinearity which may be in saturation with near zero gradient so truncating the error to zero. For instance ReLU [8] has zero gradient if the signal is zero or less, and sigmoid or tanh have nearly zero gradient as the forward signal approaches 1. The problem only arises if a network or sub-network is significantly deep, and arguably all the most successful modern networks have multiple depths using skips and lateral connections to *bootstrap* the training so alleviating the vanishing gradients problem that plagued earlier linear designs, and in particular the prevalent use of ReLU avoids the forward saturation problem entirely.

The covariate shift problem is more complex and arises during training as the model shifts often rapidly causing downstream parameter training chaos. A philosophical explanation is offered here to clarify this problem. It is conjectured that during training the network finds local minima for each neuron based upon the upstream forward signal and downstream backward error signal but critically the backpropagation algorithm makes a local decision of gradient descent without knowledge of the future shift of the model as a whole and this leads to the problem that change upstream can be more rapid than a particular neuron can deal with effectively given a constant learning rate applied to all neurons equally and so the neuron is not able to "keep up" and learn effectively. Note that a weight in a filter only updates if the neuron fires and if the corresponding input map has significant response, i.e. it is gated learning. So it is quite possible for some weights in a filter to update weekly or not at all while other weights update strongly, but critically the bias and any scaling of the filter output feel the full force of both the forward signal and backward error as they are updated on all non-zero weighted input maps. The problem of "keeping up" lag occasionally leads to a cascade tipping of local minima propagated through the net resulting in spikes in error and even the *exploding gradients* problem that leads to numerical failure. It has been often observed that the validation and training error drop suddenly after a training spike and this is the insight behind the conjecture of the cascade in local minima shifting to an upstream significant model shift trigger.

Batch normalisation (BN) has gained great popularity and has become the *de facto* standard choice for training deep convolutional neural networks. It relies upon the statistics of a small set of samples upon which a single training step is based i.e. the minibatch. BN begins with a per channel normalisation (*whitening*) of the signal based upon the statistics of the minibatch as a proxy for the statistics of the whole training dataset, and so each filter has the minibatch mean subtracted followed by division by the standard deviation. For a minibatch size of 1 this is identical to response normalisation [9]. The final stage of BN is to multiply by a trainable scaling factor **ϒ** and then add the bias **β** where (**ϒ ,β**) is the nomenclature used in the original paper referred to as the *scale* and *shift*. It is argued that they restore the ability of the batch normalised filter to represent the identity transform, and that the whitening counters both the covariate shift [6] and vanishing gradient [5] problems and thereby a network with batch normalisation trains more rapidly.

However, using minibatch statistics significantly complicates the gradient computation as the mean and standard deviation are dependent upon all samples in the minibatch and this results in increased memory footprint and significant computational overhead, and for recurrent neural networks such techniques are not applicable.

Since a deep convolutional neural network is a feedback network during training then it is argued here that what is needed instead is a built-in automatic gain control. Control theory deals with instability in feedback networks to avoid unstable dynamics. In particular automatic gain control (AGC) deals with the current state and not statistics of groups of sample points and typically seeks only to suppress the background noise level for instance using scaled mean-subtraction while amplifying the foreground signal for instance by scaling of the mean subtracted difference in the simplest case. Some parallels can be drawn with training a deep neural network, and this is the basis for the proposed approach.

The proposed neural AGC seeks to reduce the background noise level for a filter and promote the signal making only the assumption of relative *sparsity* in the signal at the beginning of training which for practical purposes means <50% signal. So mean-subtraction improves the signal to noise ratio (SNR) particularly at the beginning of training when the signal is approximately 50% foreground and less so as the model matures and the background signal is naturally reduced in level as the filters learn. Whereas BN maintains a constant mean subtraction AGC has trainable scaling of this adapting to rapid upstream model shift that introduces mean shift that temporarily destabilises the downstream net.

It is argued here that explicit minibatches are not needed as a regulariser since the gradient momentum is an implicit *recursive* minibatch. For learning rate L a parameter with gradient $g_i$ at training step **i** has gradient momentum $G_m = L.g_i + m.(G_{i-1})$ which is a recursive filtering of the gradient , so unwrapping gives an infinite time weighted minibatch where each sample has a slightly different model :

$$G_m = L.g_i + m.( L.g_{i-1} + m.( L.g_{i-2} \ldots)) = L.( g_i + m.g_{i-1} + m^2..g_{i-2} + m^3.g_{i-2} + \ldots)$$

It is proposed here that the two important elements for dealing with covariate shift are the per sample scaled mean-subtraction combined with a simple scaling of the result. The scaled mean subtraction is proposed to adaptively deal with *mean shift* [13] in the upstream signal while the *scaling* rapidly deals with *amplitude shift*. Since both the mean scaling and resulting difference scaling apply to the filter output signal then it is argued that these parameters train far more rapidly because they respond to all weighted input signals compared to individual weights within the filter that respond to their corresponding single input map which may be very sparse.

The problem of covariate shift is further exacerbated by what here is termed *semantic effective minibatch size* (SEMS) and *gated effective minibatch size* (GEMS) particularly at deeper layers which have more specific and sparser response. The problem arises because as the input signal is successively transformed to progressively higher levels of abstraction (ie more specific) it necessarily becomes more sparse since the semantic content of the input that excites a particular deep neuron is increasingly specific and less likely to be present in an input sample or even minibatch, and at the extreme may relate to a particular whole object, so what happens if that object class is not present in any sample within the minibatch? The gradient normalisation process for a convolutional filter simply sums the gradients at that parameter over the map dimensions normalised to the total number of map points. The result is that the gradient scales with the semantic content of the minibatch or sample to which it is trained, so for deeper layers then effective minibatch size becomes greatly reduced but critically the gradient is still normalised to the *entire number of points in the minibatch* for the map size, and so deeper neurons that have sparser response have attenuated gradients simply due to the naïve normalisation and so learn more slowly and are also less able to keep up with upstream more rapid model learning.

To counter these problems firstly it is proposed to instead normalise the gradients not by the mapsize*minibatchsize but to the number of non-saturated points in the map or minibatch thereof and this is termed *gated effective minibatch size* (GEMS). Secondly it is proposed to acknowledge that the deeper more abstract response maps have gradients that *superpose* rather than adding so increasing minibatch size effectively reduces their learning rate which must be restored by simply *multiplying the base single sample learning rate by the minibatch size*. Since shallower neurons are less specific then they are more likely to have significant response to all input samples, and it is observed that the increased learning rate is not destructive to their training. The GEMS work is at an early phase and is not presented further here but shows promise.

It is conjectured that the division by the standard deviation in BN has one effect of countering the naïve normalisation of the gradient for sparse maps by up-scaling the signal and hence gradient for the next layer's weights.

It is noted that the regularising effect of a minibatch proposed by BN [2] as a data augmentation already exists in the gradient momentum term and identical training speed and accuracy is achieved with single sample gradient descent. Also it is conjectured that the whitening in BN is an implicit deep data augmentation that hardens filters to amplified noise levels acting as a path *drop in* that deeply assaults the filters with noise as a function of the minibatch statistics, and this suggests a technique for infinitely varying data augmentation by deep layer-wise noise injection independent of the dataset or minibatch size. This will be reported separately and indeed does reduce over-training and improves final validation error.

The main contributions of this paper are:
1. A new simple single sample based automatic gain control technique that at least matches the accuracy of batch normalisation but with much lower memory footprint and greatly reduced computational cost.
2. An insight into the role of effective minibatch size that proposes that the training minibatch size is not a constant and one simple rule-of-thumb is to multiply the base single sample learning rate by the minibatch size.
3. Single sample training is as effective as minibatch based.
4. Insight into the operation of batch normalisation and an alternative explanation is given that fits the facts observed.

## 2 Deep automatic gain control

The proposed automatic gain control for a particular filter is formulated as follows omitting any filter index subscripts for clarity so this relates to a single filter within a convolutional layer and, where compatible, the nomenclature of batch normalisation is adopted:

$$O_s = ( I_s * W - \lambda . \overline{I_s * W} ) . \gamma + \beta \tag{1}$$

where $O_s$ is the neuron output for input maps $I_s$ for sample index $s$ in the minibatch, $W$ is the weights tensor for the filter, $I_s * W$ is the convolution of the filter weights tensor $W$ with the input $I_s$ across all map positions, $\lambda$ is the trainable scaling coefficient for the mean subtraction, $\overline{I_s * W}$ is the mean of the convolution output for input index $s$ (i.e. per sample mean not minibatch), $\gamma$ is the scaling coefficient for the mean-subtracted signal, and $\beta$ is the bias for the filter. Note that in this formulation ($\lambda$, $\gamma$, $\beta$) are all single scalars since only a single filter is considered. Gradients may then be computed on minibatches as normal.

On first glance this bares similarity to batch normalisation. However, whereas batch normalisation computes the mean across the entire minibatch the AGC computes the per sample mean, the difference is not normalised to the standard deviation and so the difference retains its absolute value, and finally the per sample filter output mean is scaled by the trainable parameter $\lambda$ and this allows the mean subtraction to be scaled down (or up) during training or restored quickly according to the demands of the optimisation. However, the difference scaling $\gamma$ is identically formulated to BN but without the level normalisation it is only tasked to deal with tracking overall signal scaling agnostic of the minibatch size.

Note however that the AGC technique could be further developed, for instance the AGC could be performed with a sliding window mean estimation across map positions or the mean could be replaced with ranking such as median or indeed any arbitrary background signal level estimation but this is a challenge with the current Theano based implementation.

Please note that the automatic gain control mechanism presented here is the subject of patent application GB1619779.0 (23rd Nov 2016) which also details the supporting explanations. Unlimited license is granted for academic and evaluation use but commercial organisations are invited to contact DeepSee AI Ltd through the website deepsee.ai for granting of license for commercial deployment of the mechanism.

## 3 Experiments and results

### 1.1 Baseline networks and experiments

The goal of training is speed without sacrificing accuracy, but a secondary goal is speed alone so long as the reduction in accuracy is minor so that new designs may be rapidly prototyped. In all experiments a fixed training rate is applied over 150 epochs. It should be noted that at 150 epochs the experiments may not have yet converged entirely and in any case no learning rate annealing is applied to fine tune. The training curves are plotted for validation results against epoch number so that the speed of training is measured in terms of number of samples seen independently of minibatch size.

Two network designs are considered with very different architectures for the demanding task of semantic segmentation for which the validation error is measured by the mean fraction of incorrectly labelled pixels. The Cityscapes [1] dataset is used throughout at resolution 256x512 with 19 classes, 2750 training samples and 500 validation samples. All results are presented in figure 1.

The two networks are described briefly below and detailed in appendix A.

   i. *Segnet* [10] : This has an encoder / decoder design each with 5 pooling layers and 13 convolutional layers with layer widths of [64, 64, 128, 128, 256, 256, 256, 512, 512, 512, 512, 512, 512] and symmetric decoder with final decoder layer width equal to the number of classes, here 19. The pooling indices version of Segnet is chosen so that the input signal must traverse the full 26 convolutional layers and has no lateral connections. This network was chosen as it is relatively deep with no skips and is familiar to most deep learning practitioners.

ii. *ClassNet* : At the time of writing this is unpublished but a link will be provided to the online paper. The network shares a similar but deeper and narrower (up to 256 filters) encoder design to the Segnet version used here and contains lateral connections preceding each pooling and has 7 pooling layers in both encoder and decoder. The lateral connections perform multi-resolution class detection and provide multiple paths and depths to the network. The down-sampled image is provided to the first layer after each pooling in the encoder providing multiple depths to the encoder. The decoder both up-samples and merges the lateral class path and jointly optimises instance boundary location and pixel class labelling losses. The lateral contour path provides deep guidance to the class up-sampling with total decoder width of typically 20 for Cityscapes using 14 classes (ie NumClasses+10). The reason for this choice of comparison network is that it compresses the decoder path and encourages dense response maps, has multiple depths from just a few layers to around 50 and has the characteristic of bootstrapping the learning by learning fine detail first and adding larger structures and context as it converges so that the effective depth at any point in training is just a few active layers learning new structure alongside many layers that are jointly fine tuning. Typically a 1x1xdepth projection is placed after every 3D convolution to support separate lateral inhibition which gives a small increase in accuracy and also a final 1x1 projection concatenates all layers in a block to support multiple depths at each resolution in the encoder and decoder which improves accuracy a little.

Both networks were trained end-to-end from a completely randomly initialised state using the fan-in variant of He [11] and momentum of 0.9 using gradient descent but also setting identity in all layers except the first after a pooling works as accurately and trains faster.

Experiments were arranged to explore the relationship between the hyper parameters of minibatch size and learning rate comparing both batch normalisation and AGC methods alongside each other for both of the chosen network designs. Due to the limitations of computational resources available, here four nVidia TitanX Pascal GPU's, then only minibatch sizes of (1, 4, 8) and learning rates of (0.02, 0.08, 0.16) were explored. The main reason for the choice of 8 for the largest minibatch size is that with Segnet and 12GB of GPU memory this is the largest minibatch size that can be trained using batch normalisation without running out of memory. The reason for the range of learning rate will be explained in the discussion, and 0.02 should be considered the *base* learning rate.

The class weights used for training were according to the ENET [12] method but it should be noted that these sum to around 30 for Cityscapes and were then normalised so that the mean class weight is 1 i.e. they sum to the number of classes. The rationale for this is that there is no apparent reason for the weights to have a mean other than 1, and that this otherwise has the effect of multiplying the learning rate by 30 for linear activations. As the goal here is to explore absolute learning rate then the procedure was adopted that class weights should always be normalised so as not to influence the absolute learning rate hyper parameter. That said, for the ClassNet variant it was decided to simply add the contour and class losses and it could be argued that this doubles the effective learning rate. In practice this was not found likely because the contour loss uses a special loss function based on the intersection over union (IoU) of only the points near the instance boundaries and acts in the role of a guide to the class segmentation, but this is only conjecture at this point.

Of interest is the value for the mean scaling $\lambda$ and in particular its final value. This has not been investigated in any great depth however it is noteworthy that the observed range is from 0 to 2 with typical mean for a mid depth layer of 0.7 to 0.9. Also there is correlation to the layer position in the net and the size of the filter. For instance near the final classifier layer $\lambda$ is close to 0 and at the deepest layers $\lambda$ typically is closer to 1 for both Segnet and ClassNet. The interpretation is that the mean shift characteristics are determined by position in the net and dealing with it is needed more in the deeper layers and less so close to the input or loss layers. A thorough examination of this phenomenon may shed some light on the behaviour of mean shift and will be pursued in follow-on work particularly by graphing $\lambda$ for particular layers throughout the entire training run.

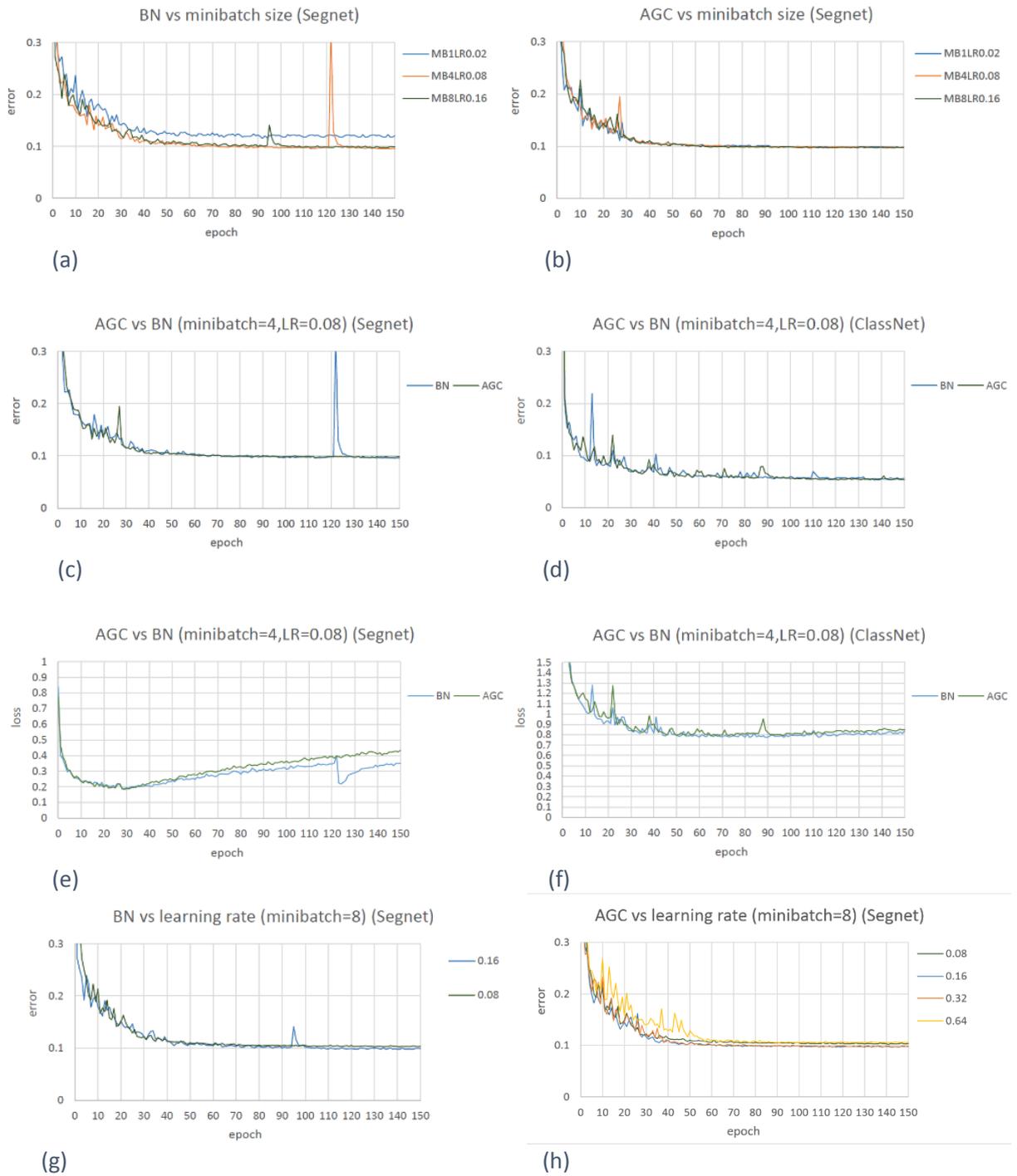

Figure 1 : Comparison of BN and AGC techniques for validation set

(a) BN error with Segnet for minibatch (1,4,8) and learning rate (0.02,0.08,0.16)
(b) AGC error with Segnet for minibatch (1,4,8) and learning rate (0.02,0.08,0.16)
(c) AGC vs BN error with Segnet at minibatch 4 and learning rate 0.08
(d) AGC vs BN error with Classnet at minibatch 4 and learning rate 0.08
(e) AGC vs BN loss with Segnet at minibatch 4 and learning rate 0.08
(f) AGC vs BN loss with Classnet at minibatch 4 and learning rate 0.08
(g) BN vs learning rate with Segnet for minibatch 8 and learning rate (0.08,0.16)
(h) AGC vs learning rate with Segnet for minibatch 8 and learning rate (0.08,0.16,0.32,0.64)

### 1.2 Learning rate VS minibatch size

Figure 1 (a) and (b) show the validation error curves for batch normalisation (BN) and AGC respectively against minibatch sizes from 1 to 8 with the corresponding learning rate set according to the base rate of 0.02 multiplied by the minibatch size. The AGC error curves show clearly that the minibatch size with scaled learning rate gives the same shape for all curves. However, with batch normalisation minibatch of 8 and 4 show similar curves but at the extreme of a single sample the training curve shows considerably higher error throughout and ends up around 2% worse which is to be expected since for single samples this is response normalisation that no longer has a proxy for batch statistics. Note also both BN and AGC converge to the same final error at the same rate. 8 is the largest minibatch size possible with Theano implementation with 256x512 input image size on a single TitanX Pascal GPU.

### 1.3 Batch normalisation VS AGC

Figure 1 (c) and (d) show the validation error curves for BN and AGC plotted against each other for Segnet and Classnet for minibatch 4 and learning rate 0.08 which is base learning rate scaled by minibatch size. Quite clearly the training curves for both techniques are almost identical though AGC outperforms by 0.2% on final error (5.34% vs 5.53%) for ClassNet and BN outperforms on Segnet by 0.16% (9.73% vs 9.57%). There are more frequent and larger training spikes for BN with both nets. The validation loss curves are also shown in (e) and (f) for completeness with slightly higher loss for the AGC technique.

### 1.4 Learning rate

Figure 1 (g) and (h) show the validation error curves for batch normalisation (BN) and AGC with Segnet for minibatch size 8 plotting learning rates of 0.16 and 0.08 against each other for comparison. Both error curves show a small and similar increase in error at the lower learning rate suggesting that the underlying mechanism and susceptibility is the same and likely the lower learning rate would catch up with longer training. AGC-Segnet was extended to learning rates of 0.32 and 0.64 to demonstrate the sweet spot.

## 4 Discussion

From the direct comparison plots of BN vs AGC it is quite clear that for the two very different styles of deep convolutional nets the training rate and accuracy for both techniques are essentially identical which supports the belief that the important elements of both techniques are the mean subtraction and the scaling $\gamma$ regardless of minibatch size so long as the learning rate is correspondingly scaled.

The very good linear correlation between minibatch size and learning rate supports the idea of semantic effective minibatch size where deeper sparser layers have gradients that superpose rather than statistically adding, and for all cases explored there is neither advantage nor disadvantage to use of different minibatch sizes in terms of absolute error or rate of training in terms of number of samples seen.

It is not clear why the validation loss for AGC is slightly higher than for BN though it is speculated that BN amplifies background noise for sparser deeper layers and acts as an implicit regulariser by augmenting the data with this noise and this is in agreement with the results presented in [2]. If this is the case then likely a technique can be developed to specifically and independently inject deep layer-wise noise to achieve the same augmentation independent of minibatch size or base training technique. This has been separately explored by the author and does indeed reduce the validation loss and will be presented elsewhere supporting this argument.

It is noted that both techniques are similarly susceptible to change in learning rate suggesting that the mechanism is likely shared and is attributed simply to the need for an optimal learning rate that is not too great, here 0.16 for minibatch of 8.

Finally, it is restated that the published learning rates are based on mean class weights of 1.0 otherwise all learning rates should be reduced typically by a factor of 30 for the class weight estimation of Segnet [11].

# 5 Conclusions

It has been demonstrated albeit for a limited range of network design and minibatch size and only with the semantic segmentation task that deep convolutional neural networks can be trained as quickly and accurately without using minibatch statistics and whitening of the convolution output and this leads to significantly simpler and faster training of the network also with the advantage of significantly lower memory footprint. For instance a 30% speed up was observed with minibatch size of 8 with a Theano implementation compared to batch normalisation with the same final error.

A strong linear correlation has been demonstrated between minibatch size and optimal learning rate which is conjectured to be based on the concept of gradient *superposition* rather than statistically adding for deeper layers with sparser response and which respond to highly specific semantic structure in the image and this leads to the concept of a *semantic effective minibatch size* which is not a constant across the layers or indeed between samples. In practical terms the base learning rate for single sample training is simply multiplied by the minibatch size to achieve the same training rate and final error.

Since the momentum term is an implicit recursive minibatch and is as effective in training with a single sample compared to a larger minibatch then it is concluded that the model change between the historical gradients is small enough not to matter and so approximate a physical minibatch of gradients that share the identical model. It is intended to further investigate this phenomenon by increasing the momentum term closer to unity.

For all practical purposes a physical minibatch appears to provide little or no advantage over single sample training with momentum both in rate of training and in final absolute error from which it is concluded that minibatches *per se* are unnecessary noting that they are in any case implicit within the gradient momentum.

This suggests a practical strategy for developing new deep networks starting initially with single sample gradient descent with low memory footprint and then fine tuning of promising designs. Of course for computational purposes minibatches provide greater efficiency on GPU implementations and result in speed up of training and may scale across multiple GPU's to reduce total training time.

Since the mean scaling parameter $\lambda$ generally maintains a significant value after convergence then it is concluded that mean shift is persistent in convergence. It is noted that shallower layers have $\lambda$ closer to zero and deeper layers appear in general to have values closer to 1 but this phenomenon has not been investigated yet systematically. Indeed some layers have slight negative values and some greater than 1.

Further work is needed to demonstrate whether the simple automatic gain control method can be as effective with much deeper networks and other tasks such as image classification and for fully connected layers but since the technique is so simple to implement (i.e. one line of code) then the reader is encouraged to just try this out with their own network designs. That said the focus of our work is practical real-time deep convolutional networks for embedded applications where the number of layers is typically more limited.

## Acknowledgements

We gratefully thank Hella Aglaia gmbh (Berlin) and in particular Dr.Oliver Klenke for their ongoing belief and financial support of this and other work and for the supply of two nVidia TitanX Pascal processors.

# References

[1] Marius Cordts, Mohamed Omran, Sebastian Ramos, Timo Rehfeld, Markus Enzweiler, Rodrigo Benenson, Uwe Franke, Stefan Roth, Bernt Schiele. *The Cityscapes Dataset for Semantic Urban Scene Understanding*, https://arxiv.org/abs/1604.01685. 2016

[2] Sergey Ioffe, Christian Szegedy. *Batch Normalization: Accelerating Deep Network Training by Reducing Internal Covariate Shift*. https://arxiv.org/abs/1502.03167v3. 2016

[3] Diederik P. Kingma, Jimmy Lei Ba. *ADAM: A Method for Stochastic Optimization*. https://arxiv.org/pdf/1412.6980. 2014

[4] Tim Salimans, Diederik P. Kingman.. *Weight Normalization: A Simple Reparameterization to Accelerate Training of Deep Neural Networks*. https://arxiv.org/pdf/1602.07868. 2016

[5] S. Hochreiter. *Untersuchungen zu dynamischen neuronalen Netzen.* Diploma thesis, Institut f. Informatik, Technische Univ. Munich. 1991

[6] Shimodaira, Hidetoshi. *Improving predictive inference under covariate shift by weighting the log-likelihood function*. Journal of Statistical Planning and Inference, 90(2):227–244, October 2000.

[7] David Rumelhart, Geoffrey Hinton, RonaldWilliams. *Learning representations by back-propagating errors*. ature. 323 (6088): 533–536. 1986.

[8] Nair, Vinod and Hinton, Geoffrey E. *Rectified linear units improve restricted boltzmann machines*. ICML, pp.807–814. Omnipress, 2010.

[9] Alex Krizhevsky, Ilya Sutskever, Geoffrey E. Hinton. *ImageNet Classification with Deep Convolutional Neural Networks*. NIPS, 2012

[11] Vijay Badrinarayanan, Alex Kendall, Roberto Cipolla. *Segnet: A Deep Convolutional Encoder-Decoder Architecture for Image Segmentation*. https://arxiv.org/abs/1511.00561. 2015

[11] He, Kaiming, Zhang, Xiangyu, Ren, Shaoqing, and Sun, Jian. *Delving deep into rectifiers: Surpassing human-level performance on imagenet classification.* International Conference on Computer Vision (ICCV), 2015

[12] Adam Paszke, Abhishek Chaurasia, Sangpil Kim, Eugenio Culurciello. *ENet: A Deep Neural Network Architecture for Real-Time Semantic Segmentation.* arXiv:1606.02147, 2016

[13] Djork-Arn´e Clevert, Thomas Unterthiner & Sepp Hochreiter. *ENet: Fast and Accurate Deep Network Learning by Exponential Linear Units (ELUS).* Proceedings ICLR 2016

# Appendix A – ClassNet architecture

## A.1 Block design

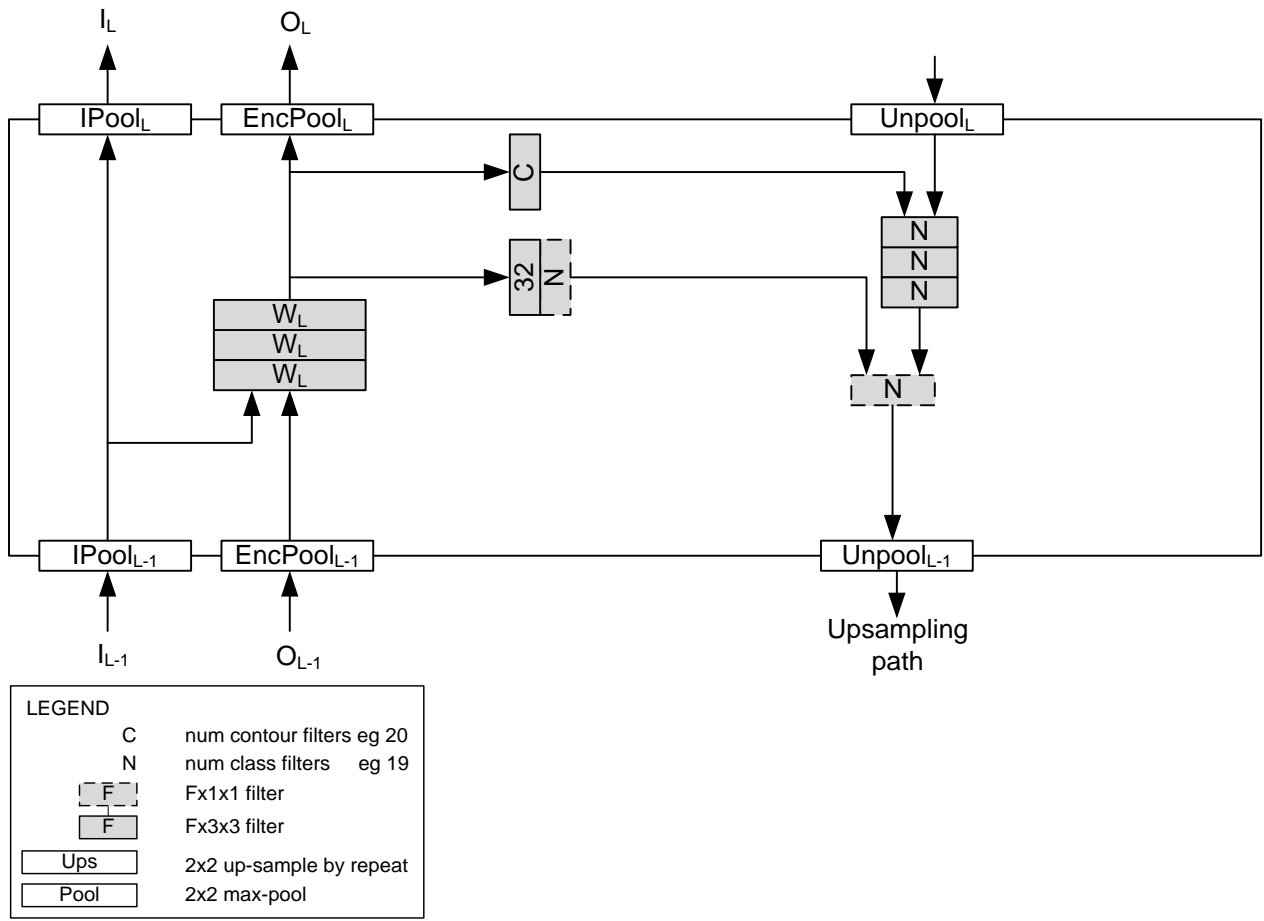

## A.2 ClassNet convolutional network overview (7 blocks)

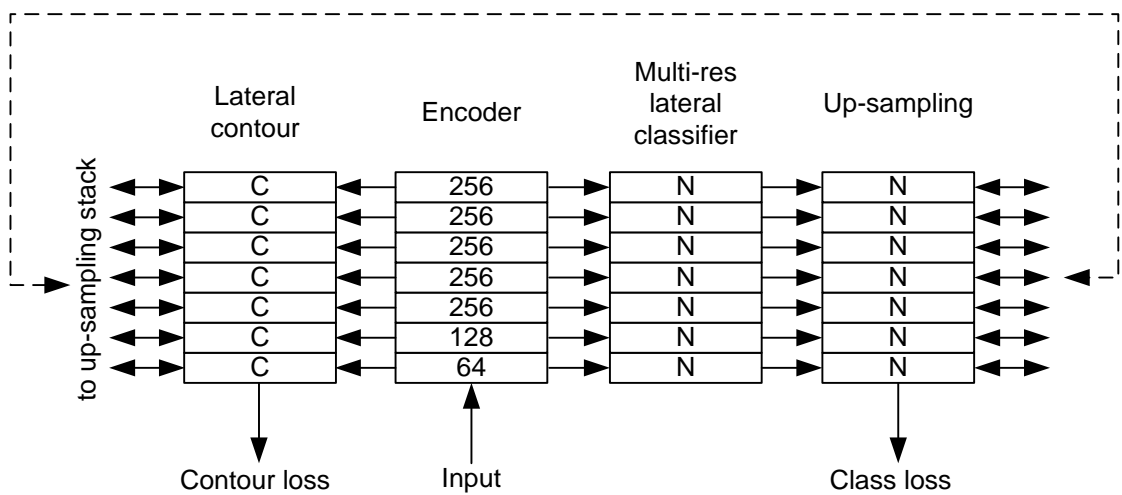